\documentclass[runningheads]{llncs}
\usepackage{graphicx}
\usepackage{graphics} 
\usepackage{epsfig} 
\usepackage{mathptmx} 
\usepackage{times} 
\usepackage{amsmath} 
\usepackage{amssymb}  
\usepackage{algorithm}
\usepackage{algpseudocode}
\usepackage{cite}
\usepackage{hyperref}
\usepackage{booktabs}
\usepackage{caption}
\usepackage{makecell} 
\usepackage{multirow}
\usepackage{adjustbox}
\usepackage{changepage}
\usepackage{xcolor}
\usepackage{wrapfig}
\usepackage{subcaption}
\DeclareMathAlphabet{\mathcal}{OMS}{cmsy}{m}{n}

\hypersetup{
  colorlinks   = true,    
  urlcolor     = blue,    
  linkcolor    = blue,    
  citecolor    = blue      
}

\begin{document}
\title{Dual Action Policy for Robust Sim-to-Real Reinforcement Learning} 
%
%

\author{Ng Wen Zheng Terence \inst{1,2}\and
Chen Jianda \inst{1}}

\authorrunning{NWZ.T et al.}

\institute{Nanyang Technological University, Singapore \and
Continental Automotive, Singapore \\
\email{ngwe0099@e.ntu.edu.sg}\\
\email{jianda001@e.ntu.edu.sg}
}

\maketitle              
\begin{abstract}

This paper presents Dual Action Policy (DAP), a novel approach to address the dynamics mismatch inherent in the sim-to-real gap of reinforcement learning. DAP uses a single policy to predict two sets of actions: one for maximizing task rewards in simulation and another specifically for domain adaptation via reward adjustments. This decoupling makes it easier to maximize the overall reward in the source domain during training. Additionally, DAP incorporates uncertainty-based exploration during training to enhance agent robustness. Experimental results demonstrate DAP's effectiveness in bridging the sim-to-real gap, outperforming baselines on challenging tasks in simulation, and further improvement is achieved by incorporating uncertainty estimation.

\keywords{Reinforcement Learning  \and Sim-to-Real \and Domain Adaptation}
\end{abstract}

\section{Introduction}

\emph{Reinforcement learning} (RL) has revolutionized the way we train intelligent agents. Conventional RL involves real-world interactions, which is expensive, time-consuming, and even dangerous. The use of simulators is emerging as a powerful alternative, offering numerous benefits for RL training \cite{lee2020learning, gu2017deep}. Simulators provide a safe, controlled, and cost-effective environment for agents to learn. Unlike the real world, where mistakes can have significant consequences, simulations allow agents to explore freely and experiment without fear of failures. 
This shift from the real world to simulated environments opens the door to training intelligent agents for various applications, paving the way for advancements in fields like robotics \cite{lee2020learning, gu2017deep}, healthcare \cite{liu2020reinforcement}, and finance \cite{charpentier2021reinforcement}. 

While simulators offer numerous advantages for training RL agents, they also introduce challenges such as \emph{sim-to-real gap} \cite{chebotar2019closing, peng2018sim}. Despite efforts to create realistic simulators, discrepancies between simulated and real-world environments can lead to performance degradation when deploying agents trained solely in simulation to the physical world. Factors such as inaccuracies in physics modeling, sensor noise, and environmental variations can all contribute to this mismatch. 
This can lead to sub-optimal performance or failure, emphasizing the need to bridge the sim-to-real gap for effective and safe RL agents.

Several approaches have been explored to tackle the sim-to-real gap, and this work specifically focuses on the challenge of \emph{dynamics mismatch}.
 One common approach is domain randomization (DR) where RL policies are trained across a diverse range of simulated dynamics \cite{peng2018sim, yu2017preparing}. While DR improves real-world performance by training adaptable policies, it often requires prior knowledge of parameter variations.
Another contrasting approach involves grounding the simulator to resemble the target domain, allowing the agent to learn as if it were directly interacting with the real world \cite{hanna2017grounded,karnan2020reinforced,Christiano2016TransferFS}. A similar strategy involves reward shaping, where a bonus reward incentivizes the agent to take actions that mimic those preferred in the real world \cite{eysenbach2020off, niu2022trust}. 
Both grounding and reward shaping techniques require some level of interaction with the target environment, which contradicts our objective of avoiding real-world interactions due to safety and feasibility concerns. 
Therefore, we focus on a more realistic approach by leveraging a small, sub-optimal offline dataset from the target environment to reduce the sim-to-real gap \cite{jiang2021simgan}. This eliminates the need for further real-world interactions, making it well-suited for real-world scenarios, e.g., operating rescue robots in hazardous environments, deploying trading strategies in high-frequency markets, or controlling unmanned aerial vehicles in challenging settings.

We propose a novel method, Dual Action Policy (DAP), which uses a single policy to predict two sets of actions simultaneously. One set maximizes task rewards in the simulation environment, while the other specifically addresses dynamics mismatch. Inspired by prior work \cite{eysenbach2020off}, we introduce a reward adjustment mechanism to incentivize actions resembling those in the target domain. By decoupling actions, the agent can optimize task rewards while adjusting for dynamics.
Furthermore, we proposed an action resampling method during training to enhance agent robustness by encouraging exploration in areas with high epistemic uncertainty. During deployment, this allows the agent to self-correct and return to more certain state-action distributions. 

Our experiments demonstrate the effectiveness of DAP for bridging the sim-to-real gap. We compared various methods on challenging simulated tasks with mismatched dynamics. DAP outperformed all baselines, achieving significantly higher returns than the strong baselines. Additionally, incorporating uncertainty estimation further improved performance, nearly matching the optimal results in some cases. Ablation studies validated the importance of the uncertainty-based exploration and showed that DAP remains effective even with a limited amount of target data.

\section{Related Work}

\subsection{Dynamics Mismatch in Sim-to-Real RL}

Several methods have been proposed to bridge the sim-to-real gap in RL, particularly the challenge of dynamics mismatch between training and deployment environments.
These methods fall into three main categories: system identification, domain randomization, and domain adaptation.
\emph{System identification}, the earliest and most established approach, uses offline data to calibrate the simulator, essentially parameterizing the simulated environment to better match the real world \cite{chebotar2019closing}.
A closely related solution is \emph{online} system identification, which directly utilizes inferred system parameters to update a meta-policy \cite{yu2017preparing}. Both methods are often data and computationally expensive, especially for complex, high-dimensional environments.
\emph{Domain randomization} (DR) offers a contrasting perspective. Here, RL policies are trained across a diverse range of simulated dynamics \cite{peng2018sim, yu2017preparing}. While DR fosters more adaptable policies that perform better in real-world settings, its effectiveness often requires a prior knowledge of parameter types and ranges.

Recent innovations include \emph{domain adaptation} strategies, exemplified by methods like reward adjustment, which penalize actions with high dynamic discrepancies during source domain training \cite{eysenbach2020off}. 
This adjustment relies on auxiliary classifiers trained separately to distinguish between source and target domain transitions. 
Another similar approach, involving reward adjustments, combines an offline dataset with online samples from simulation \cite{niu2022trust}. However, this method assumes the offline dataset consists of near-optimal samples in order to perform effectively.
Lastly, a hybrid category combines elements of system identification and domain adaptation. This line of work, exemplified by \emph{grounded} simulators \cite{hanna2017grounded,karnan2020reinforced,Christiano2016TransferFS}, employs transformations on the source domain to align its behavior with the target domain. This essentially allows the agent to learn as if it were directly interacting with the target environment. These transformations incorporate forward and/or inverse dynamics models of both domains, enabling agents to learn within the context of the target domain. One limitation of this approach is that the performance is highly dependent on the state-action coverage achieved by the grounding policy.

\subsection{Uncertainty in Deep Reinforcement Learning}

Deep Neural Networks (DNNs) have been instrumental in the success of deep reinforcement learning (DRL) due to their ability to represent complex functions \cite{mnih2015human}, including smooth dynamics often present in robotics \cite{lillicrap2015continuous}. However, when data is limited, DNNs are susceptible to over-fitting, which can lead to significant uncertainty about the model's true capabilities and degrade the performance of deep RL frameworks. A systematic approach to address this issue is parametric Bayesian inference \cite{blundell2015weight}. This method leverages a predefined probability distribution to represent the uncertainty in a model's parameters. However, its performance heavily relies on choosing an informative prior, and calculating the posterior distribution in complex models with many parameters, which can still be computationally expensive, especially for high-dimensional data.

An effective alternative is to use non-parametric bootstrapping. This technique leverages an ensemble of models with random initializations. Bootstrapping ensembles have found applications in various areas of DRLs. For example, they are used for learning an accurate dynamics model in model-based RL \cite{janner2019trust, chua2018deep}. They have also been used in policy and value functions within model-free RL \cite{kurutach2018model, osband2016deep} and offline RL \cite{ghasemipour2022so, an2021uncertainty}. Furthermore, model uncertainty estimation plays a crucial role in designing exploration rewards \cite{osband2016deep, liang2021reward}.  In our work, we leverage bootstrapped models to estimate uncertainty in the dynamics modelling, a factor we demonstrate to be crucial for domain adaptation.

\section{Background}

\subsection{Reinforcement Learning}

In RL \cite{sutton2018reinforcement}, an environment is characterized by a Markov Decision Process (MDP) denoted as $\mathcal{M} = (S, A, P, R, \gamma, d_0)$. Here, $S$ and $A$ represent the state and action spaces, respectively. The transition dynamics are defined by $P: S \times A \times S \rightarrow [0,1]$, and the reward function is represented by $R: S \times A \times S \rightarrow \mathbb{R}$. The discount factor is a scalar $\gamma \in [0, 1)$ that reflects the agent's preference for immediate rewards over future rewards. Finally, $d_0$ defines the initial state distribution. 
An agent's policy, denoted by $\pi: S \rightarrow A$, induces a probability distribution over trajectories, $\tau = (s_t, a_t, r_t)_{t\geq0}$. Here, $s_t$, $a_t$, and $r_t$ represent the state, action, and reward at timestep $t$, respectively. 
The goal of Maximum Entropy RL is to find the optimal policy, $\pi^*$, that maximizes expected cumulative discounted reward and a entropy-regularized term: $\pi^* = \arg\max_{\pi} \mathbb{E}_{\tau \sim p_{\pi}}[\sum_{t}{\gamma^t (r_t + \mathcal{H}_\pi\left[a_t \mid s_t\right])}]$, where $\mathcal{H}_\pi$ is entropy of the policy \cite{haarnoja2017reinforcement}.

\subsubsection{\textbf{Problem Statement.}}
We intend to deploy our agent in a given environment $\mathcal{M}_\text{target} = (S, A, P, R, \gamma, d_0)$. However, due to the constraints on interacting directly with the target environment, we leverage interactions with a more accessible simulation environment, $\mathcal{M}_\text{source} = (S, A, P', R, \gamma, d_0)$. The key difference between these two environments lies in their transition dynamics, i.e. $P \neq P'$. In our problem setting, we assume abundant interactions with $\mathcal{M}_\text{source}$ and a limited and sub-optimal dataset collected from $\mathcal{M}_\text{target}$.

\subsection{Domain Adaptation with Rewards from Classifier (DARC)}
To mitigate the dynamics mismatch, Eysenbach et al. \cite{eysenbach2020off} proposed a domain adaptation method DARC, which learns a policy whose behavior receives high reward in the source domain and has high likelihood under the target domain dynamics. Specifically, they minimized the reverse KL divergence between the desired distribution over trajectories in the target domain $p(\tau)$ and the agent’s distribution over trajectories in the source domain $q(\tau)$ as follows:
\begin{equation}
\label{eqn:darc}
    \min _{\pi(a \mid s)} D_{\mathrm{KL}}(q \| p)=-\mathbb{E}_{p_{\text {source }}}\left[\sum_t r\left(s_t, a_t\right)+\mathcal{H}_\pi\left[a_t \mid s_t\right]+\Delta r\left(s_t, a_t, s_{t+1}\right)\right]
\end{equation}
where 
$$\Delta r\left(s_t, a_t, s_{t+1}\right) \triangleq \log p\left(s_{t+1} \mid s_t, a_t\right)-\log q\left(s_{t+1} \mid s_t, a_t\right).$$
$\mathcal{H}_\pi$ is the entropy of the policy and $t$ is the time-step. The reward adjustment $\Delta r$, penalises the agent for taking transitions more likely in the source domain than in the target domain and vice versa. 
The reward adjustment $\Delta r$  requires an explicit model of the dynamics which may be inaccurate in continuous control tasks with a high dimensional state-action space. Subsequently, they estimated $\Delta r$ using Bayes rules with two domain classifiers $p(\cdot | s_t, a_t, s_{t+1}) $ and $p(\cdot | s_t, a_t) $ as follows:
\begin{equation}
\begin{aligned}
\Delta r\left(s_t, a_t, s_{t+1}\right) & =\log p\left(\operatorname{target} \mid s_t, a_t, s_{t+1}\right)-\log p\left(\operatorname{target} \mid s_t, a_t\right) \\
& -\log p\left(\text { source } \mid s_t, a_t, s_{t+1}\right)+\log p\left(\text { source } \mid s_t, a_t\right)
\end{aligned}
\label{eqn:darc_class}
\end{equation}
The domain classifiers are binary classifiers trained separately used to distinguish between source and target domain transitions.

\section{Methodology}

The key to DARC for domain adaptation lies in the introduced reward adjustment. This adjustment incentivizes agents to choose actions that lead to transitions resembling those in the target domain. However, this approach has two limitations:

\begin{enumerate}
    \item \textbf{Action Set Constraint}: During training, the action set used to compute the reward adjustments (Equation \ref{eqn:darc}) coincides with the action set used for updating the maximum entropy RL method through simulation sampling. Restricting both actions to the same set hinders the search for an optimal solution that maximises both rewards.
    \item \textbf{Classifier Errors}: Due to the epistemic errors in the DARC classifiers, the dynamics reward adjustments might be inaccurate. These errors are particularly problematic if they are overly optimistic. In such cases, the policy might be led to sample state-action pairs that fall outside the target distribution during rollouts in the target environment. Consequently, the agent lacks the capability to self-correct and return to the desired distribution, causing planning in the target environment to diverge.
\end{enumerate}
In the following, we propose novel solutions to address these limitations.

\subsection{Dual Action Policy (DAP)}
\label{sect:dap}

\begin{figure}[t]
    \centering
    \includegraphics[width=1.1\linewidth]{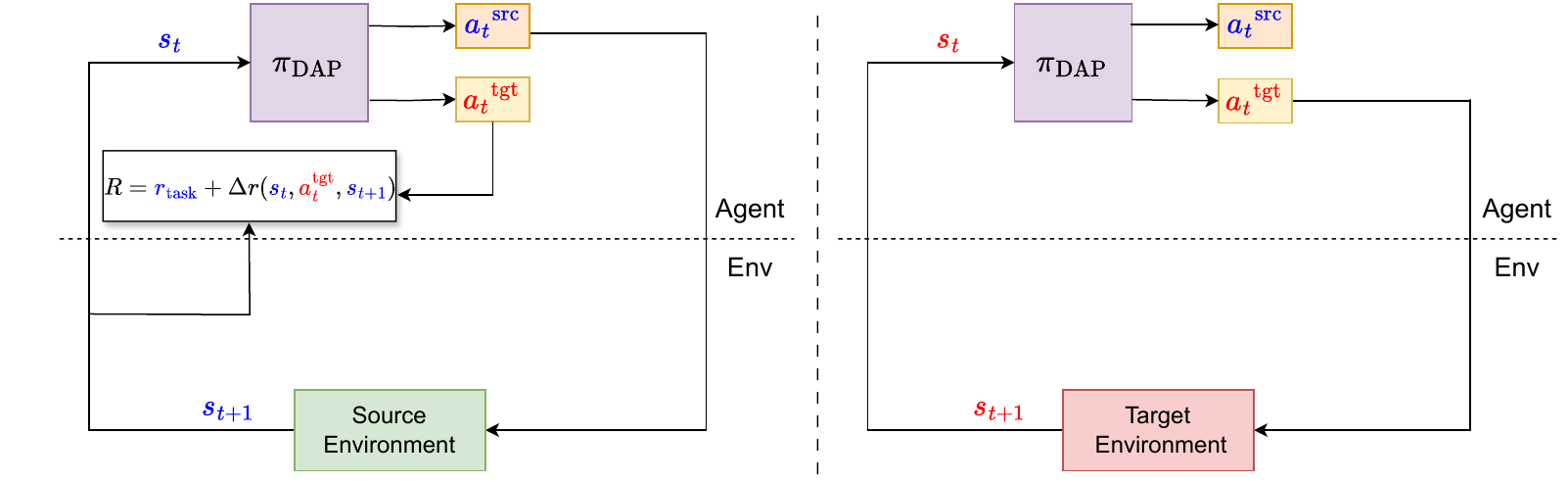}
    \caption{ (Left) Training in Source Env: $\pi_\text{DAP}$ predicts additional action set $a^\text{tgt}$, optimized for the target environment while $a^\text{src}$ is utilized for sampling the source environment. (Right) Deployment in Target Env: only $a^\text{tgt}$ is utilized while $a^\text{src}$ is discarded.
    }
    
    \label{fig:dap}
\end{figure}

The first limitation of DARC lies in the action set constraint. To address this issue, we propose a relaxation strategy using DAP. The core idea of DAP is to utilize a single policy to simultaneously predict two distinct sets of actions $a = [a^\text{src}$, $a^\text{tgt}]$. The first set $a^\text{src}$, is used for sampling within the simulation environment which aligns with the standard behavior of maximum entropy RL methods.
The second set introduces a novel concept: predicting an additional set of actions $a^\text{src}$. 
The decoupling of the actions into two sets would make it easier for $a^\text{tgt}$ to address the dynamics mismatch via reward adjustments, while $a^\text{src}$ focuses on maximizing the task reward. 
During deployment, we  utilize $a^\text{tgt}$ to sample the target environment. 
An overview of the training and deployment process for DAP is illustrated in Figure~\ref{fig:dap}. 
We highlighted all terms related to the $a^\text{src}$ or $a^\text{tgt}$ in \textcolor{blue}{blue} and \textcolor{red}{red} respectively.

Formally, following Equation \ref{eqn:darc}, the modified objective function can be expressed as:

\begin{align*}
    \begin{split}
        \min_{\pi([\textcolor{blue}{a^\text{src}_t}, \textcolor{red}{a^\text{tgt}_t}] \mid s)} D_{\mathrm{KL}}(q \| p)  = & -\mathbb{E}_{p_{\text {source }}}\left[\sum_t r\left(\textcolor{blue}{s_t}, \textcolor{blue}{a^\text{src}_t}\right)+\mathcal{H}_\pi\left(\textcolor{blue}{a^\text{src}_t} \mid \textcolor{blue}{s_t}\right) +\Delta r\left(\textcolor{blue}{s_t},  \textcolor{red}{a^\text{tgt}_t}, \textcolor{blue}{s_{t+1}}\right)       \right]
    \end{split}    
\end{align*}
and the modified reward adjustment is:
\begin{align*}
\hat{\Delta} r\left(\textcolor{blue}{s_t},  \textcolor{red}{a^\text{tgt}_t}, \textcolor{blue}{s_{t+1}}\right) 
 &\triangleq \log p\left(\textcolor{blue}{s_{t+1}} \mid \textcolor{blue}{s_t}, \textcolor{red}{a^\text{tgt}_t}\right)-\log q\left(\textcolor{blue}{s_{t+1}} \mid \textcolor{blue}{s_t}, \textcolor{red}{a^\text{tgt}_t}\right) 
\end{align*}  
This objective is optimized under a new MDP, $\mathcal{M}_\text{dual} = (S, A_\text{dual}, P, R, \gamma, d_0)$, where $|A_\text{dual}| = 2 \times |A|$, and $A$ is the original action-space. Similar to DARC, following Equation \ref{eqn:darc_class}, we use a pair of domain classifiers to estimate $\hat{\Delta} r$:

\begin{equation}
\begin{aligned}
\hat{\Delta} r\left(\textcolor{blue}{s_t}, \textcolor{red}{a^\text{tgt}_t}, \textcolor{blue}{s_{t+1}}\right) =  &\log p\left(\operatorname{target} \mid \textcolor{blue}{s_t}, \textcolor{red}{a^\text{tgt}_t}, \textcolor{blue}{s_{t+1}}\right)-\log p\left(\operatorname{target} \mid \textcolor{blue}{s_t}, \textcolor{red}{a^\text{tgt}_t}\right) \\
& -\log p\left(\text { source } \mid \textcolor{blue}{s_t}, \textcolor{red}{a^\text{tgt}_t}, \textcolor{blue}{s_{t+1}}\right)+\log p\left(\text { source } \mid \textcolor{blue}{s_t}, \textcolor{red}{a^\text{tgt}_t}\right) 
\end{aligned}
\label{eqn:dap_pre}
\end{equation}

\subsubsection{Regularization}

From our DAP formulation in Equation \ref{eqn:dap_pre}, the source and target policies play distinct but interconnected roles. The source policy faces a dual optimization challenge: it must maximize returns while generating state sequences indistinguishable from the target tasks by the classifiers. In contrast, the target policy's primary influence is on the reward shaping term. This design creates an interesting dynamic. To this end, we introduce a regularization term, $||a^\text{src}_t-a^\text{tgt}_t||_2^2$, to prevent the generation of infeasible actions by $a^\text{tgt}$, controlled by a hyperparameter $\lambda$ as follows:
\begin{equation}
\begin{aligned}
\Delta r\left(\textcolor{blue}{s_t}, \textcolor{red}{a^\text{tgt}_t}, \textcolor{blue}{s_{t+1}}\right) =  \hat{\Delta} r\left(\textcolor{blue}{s_t}, \textcolor{red}{a^\text{tgt}_t}, \textcolor{blue}{s_{t+1}}\right) +\lambda ||\textcolor{blue}{a^\text{src}_t}-\textcolor{red}{a^\text{tgt}_t}||_2^2
\end{aligned}
\label{eqn:dap}
\end{equation}

The regularization hyper-parameter $\lambda$ plays a pivotal role in maintaining the balance between these two policies. Without regularization ($\lambda\rightarrow 0$), the target policy would be unconstrained and might prioritize generating actions that maximize the reward shaping term, without necessarily optimizing for actual task returns. Conversely, large $\lambda$ values would cause DAP to converge towards DARC-like behavior, as the regularization term would dominate the reward shaping, effectively forcing the two policies to be nearly identical.

Another factor contributing to DAP's effectiveness is the simultaneous prediction of $a^\text{src}$ and $a^\text{tgt}$ using a single policy. 
Maximizing Equation \ref{eqn:dap} with a single policy constrains the target policy to ensure that $a^\text{tgt}$ stays within the state distribution reachable by $a^\text{src}$, which would be challenging if separate models were used.

\subsection{Uncertainty-based Robust Action Resampling}
\label{sect:robust}
Next, we address the problem of epistemic errors in the DARC's domain classifiers, which can lead to overly optimistic strategies during deployment. Our solution tackles this issue within the training framework by first measuring these uncertainties. Thereafter, we modify the predicted action to a more robust choice based on the severity of the uncertainty. This is achieved by randomly perturbing the action with a magnitude proportional to the uncertainty level. This approach allows actions with low uncertainty to remain unchanged, while forcing uncertain actions to explore a wider range of states. By encouraging exploration in uncertain areas, the agent gains the capability to self-correct and return to state-action distributions with greater certainty.

Formally, to quantify the dynamics uncertainty due to epistemic errors, we follow a simple but effective method in \cite{lakshminarayanan2017simple} which utilizes a deep ensemble. Specifically, we train an ensemble of $N$ domain classifiers (Equation \ref{eqn:dap}), denoted by $p_i(\text{target} | s_t, a_t, s_{t+1})$ and $p_i(\text{source} | s_t, a_t, s_{t+1})$, respectively, for $i = 1, \dots ,N$, with randomly initialized weights. The intuition behind this approach is that a high standard deviation in the log probabilities across the ensemble indicates significant disagreement about the predicted state transitions, suggesting higher uncertainty in the dynamics model. During training, for each sampled action $a = [a^\text{src}$, $a^\text{tgt}]$, we resample and replace $a^\text{src}$ with a robust action, denoted as $\hat{a}_t^\text{src}$, from a normal distribution: 
\begin{equation}
\hat{a}_t^\text{src} \sim \mathcal{N}(a_t^\text{src}, k \sigma_t),
\label{eqn:robust}
\end{equation}
where
\begin{equation*}
\begin{aligned}
\sigma_t & = \text{Std} \Big\{ \left[ \log p_i\left(\text{target} \mid s_{t-1}, a_{t-1}^\text{tgt}, s_{t}\right)-\log p_i\left(\text{target} \mid s_{t-1}, a_{t-1}^\text{tgt}\right) \right.\\
& \left. \quad -\left.\log p_i\left(\text{source} \mid s_{t-1}, a_{t-1}^\text{tgt}, s_{t}\right)+\log p_i\left(\text{source} \mid s_{t-1}, a_{t-1}^\text{tgt}\right)\right] _{i=1,\dots,N} \right\}
\end{aligned}
\end{equation*}
Here, $k$ represents a scaling hyper-parameter, $\sigma_{t}$ represents an uncertainty measure
and $\text{Std}()$ is the standard deviation function. 
The modified action encourages exploration where the agent may behave erroneously in the target environment, enhancing its robustness.
During deployment, the agent directly relies on the sampled action from the policy, bypassing the robust action. The complete training procedure combining DAP and uncertainty-based action resampling is detailed in Algorithm \ref{algo:1}.


\begin{algorithm}[t]
\small
\caption{Dual Action Policies (DAP) with Uncertainty-based Action Resampling}
    \begin{algorithmic}[1]
        \State \textbf{Input:} Target dataset $\mathcal{D}_\text{target}$, Source MDP $\mathcal{M}_\text{source}$
        \State \textbf{Input:} Regularizer $\lambda$, scaling parameter $k$, ensemble size $N$.
        \State \textbf{Initialize:} $\pi_\text{DAP}$, $\mathcal{M}_\text{dual}$ using $\mathcal{M}_\text{source}$, initial state $s_1$, replay buffer $\mathcal{D}_\text{source}$ and ensemble of domain classifiers $p_i(\cdot | s_t, a_t, s_{t+1}) $ and $p_i(\cdot | s_t, a_t)$ for $i=1,\cdots,N$. 
        
        \For{$t$ in $1, 2, ..., \text{num\_iter}$}
            \State Sample $a = [a^\text{src}$, $a^\text{tgt}]$ from $\pi_\text{DAP} (s_t)$
            \State Resample $a^\text{src}$  based on Eqn. \ref{eqn:robust} to get $\hat{a}^\text{src}$
            \State Sample $s_{t+1}$ in MDP $\mathcal{M}_\text{dual}$ using $\hat{a}^\text{src}$
            \State Compute reward adjustment $\Delta r(s_t,a^\text{tgt}_t, s_{t+1}) $ using Eqn \ref{eqn:dap}
            \State Store $(s_t, a^\text{src}$, $a^\text{tgt}, r_t + \Delta r, s_{t+1})$ in $\mathcal{D}_\text{source}$
            \State Update $\pi_\text{DAP}$ with $\mathcal{D}_\text{source}$ using SAC \cite{haarnoja2018soft}
            \State Update domain classifiers with $\mathcal{D}_\text{source} \cup \mathcal{D}_\text{target}$ \cite{eysenbach2020off}
        \EndFor
    \State \textbf{return} $\pi_\text{DAP}$
    \end{algorithmic}
    \label{algo:1}
\end{algorithm}

\section{Experiments}


\subsection{Experimental Setup}

\noindent\textbf{Environments.}

To evaluate our method's ability to bridge the sim-to-real gap, we conducted experiments in the MuJoCo physics simulator \cite{todorov2012mujoco}. We used a diverse set of four challenging settings created by modifying the physical properties of simulated robots in these environments: Ant, Half-Cheetah, Hopper, and Walker2d. 
We collected an offline dataset consisting of $M = 20000$ samples using a source behavioral policy sampled in the target environment. The source behavioral policy is trained using Soft-Actor Critic (SAC) in the source environment with 1M steps \cite{haarnoja2018soft}. 

\noindent\textbf{Baselines.} We benchmarked against various sim-to-real algorithms which aim to address the dynamics mismatch. We have excluded methods which requires dynamics prior \cite{jiang2021simgan,chebotar2019closing, peng2018sim}. Overall, we trained the following policies:
\begin{enumerate}
    \item \textbf{RL on source} - Policy trained in the source environment using SAC~\cite{haarnoja2018soft}. This is also the behavioral policy used to collect the target dataset. 
    \item \textbf{RL on target} - Policy trained in the target environment using SAC~\cite{haarnoja2018soft}. \emph{Oracle} for the upper bound performance in the target environment. 
    \item \textbf{GARAT}~\cite{desai2020imitation} - Method to ground the simulator using adversarial policy.
    \item \textbf{H2O}~\cite{niu2022trust} - Dynamics-aware method to learn from both online and offline data.
    \item \textbf{DARC}~\cite{eysenbach2020off} - Method based on reward adjustments with domain classifiers .
    \item \textbf{DAP (Ours)} - Dual action policy predicting two set of actions (Sect. \ref{sect:dap})
    \item \textbf{DAP+U (Ours)} - DAP + \textbf{U}ncertainty-based Robust Action Resampling (Sect. \ref{sect:robust})
    \end{enumerate}
\noindent\textbf{Implementation details.}
We adopted the default hyperparameter settings from DARC \cite{eysenbach2020off} for our DAP implementation. All policies were trained for 1M steps. For the regularization term $\lambda$ (Equation \ref{eqn:dap}), we observed stability across values in the range $[0.05, 0.2]$, thus we fixed $\lambda=0.10$ for all experiments, except for the ablation study of $\lambda$. The number of ensembles used to calculate uncertainty was set to $N = 5$ to achieve a good balance between accuracy and speed. The scaling parameter for uncertainty-based action resampling (Equation \ref{eqn:robust}) was set to $k = 0.10$ for all experiments except for the ablation study of $k$.

\subsection{Main Results}

\begin{figure}[t]
    \centering
    \begin{subfigure}{0.49\textwidth}
        \centering
        \includegraphics[width=\linewidth]{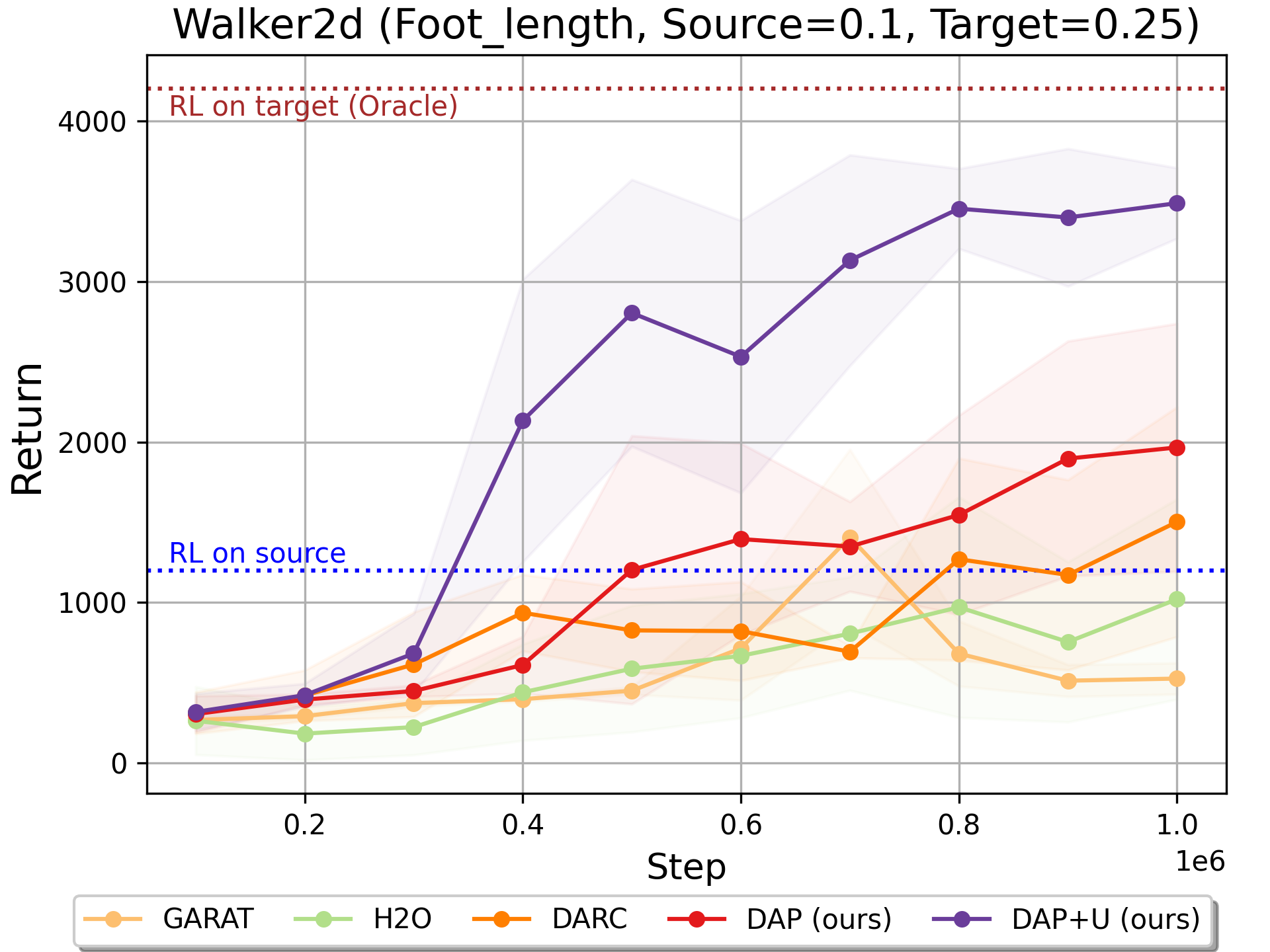}
    \end{subfigure}
    \begin{subfigure}{0.49\textwidth}
        \centering
        \includegraphics[width=\linewidth]{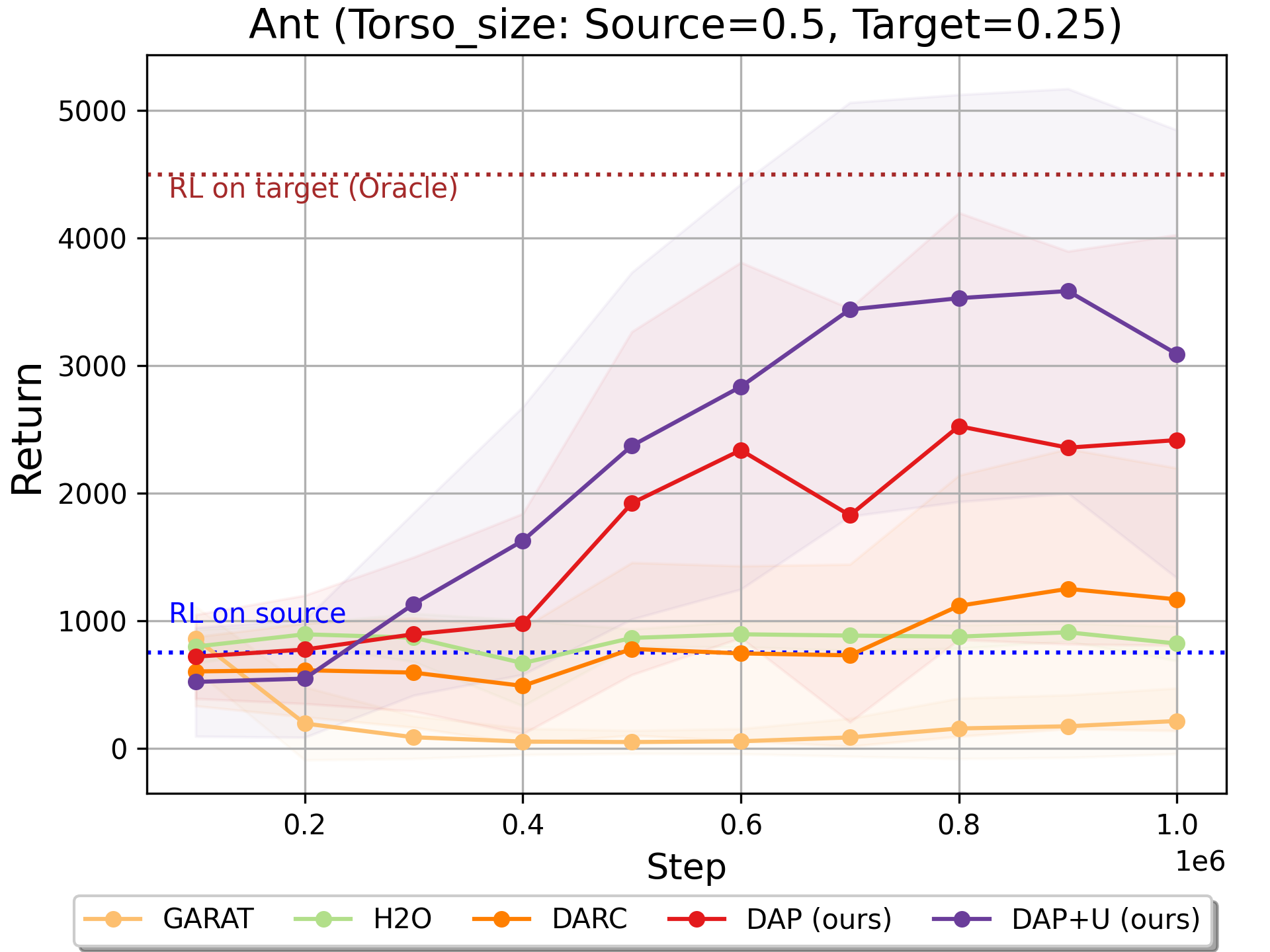}
    \end{subfigure}
    \begin{subfigure}{0.49\textwidth}
        \centering
        \includegraphics[width=\linewidth]{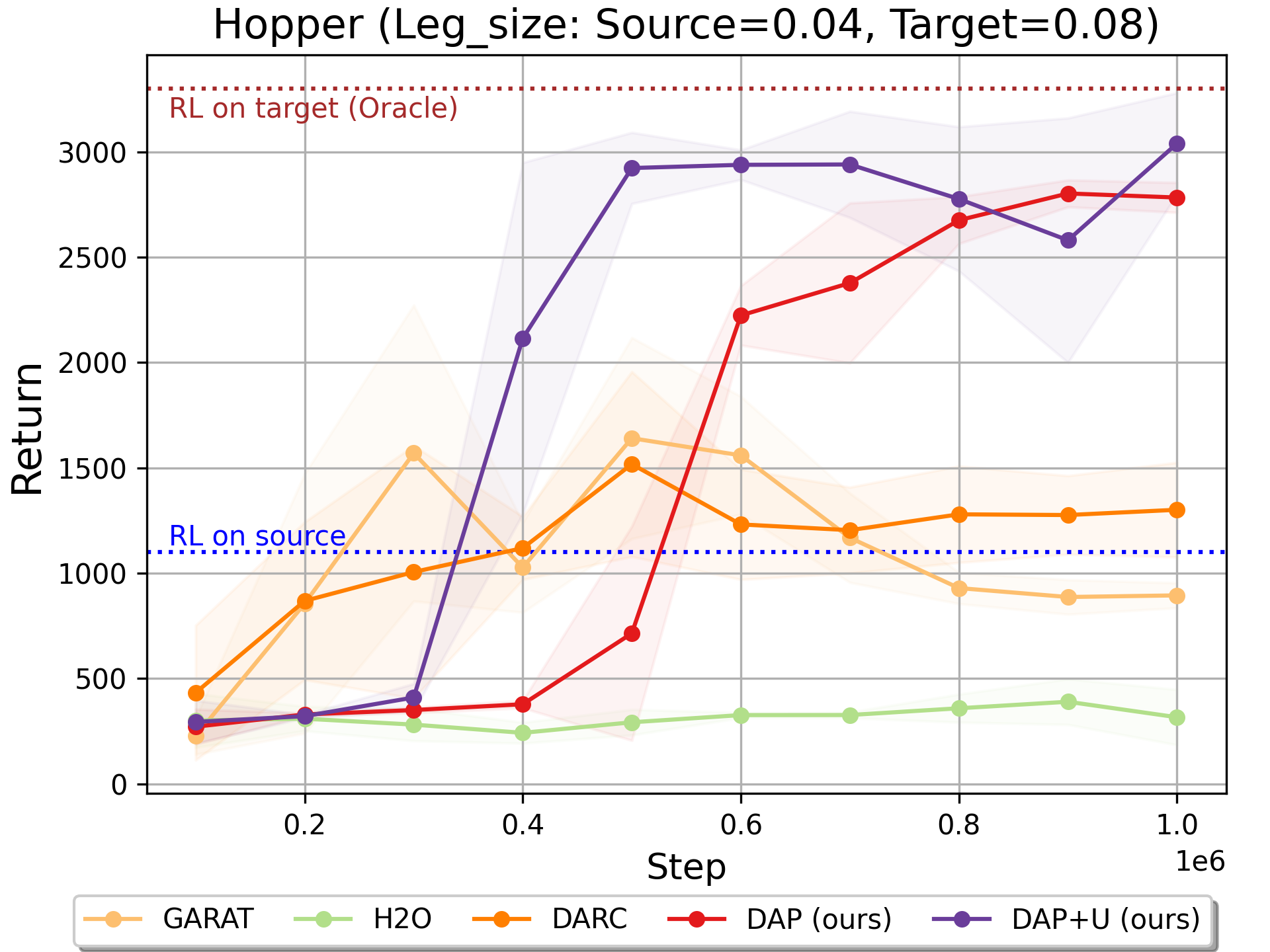}
    \end{subfigure}
    \begin{subfigure}{0.49\textwidth}
        \centering
        \includegraphics[width=\linewidth]{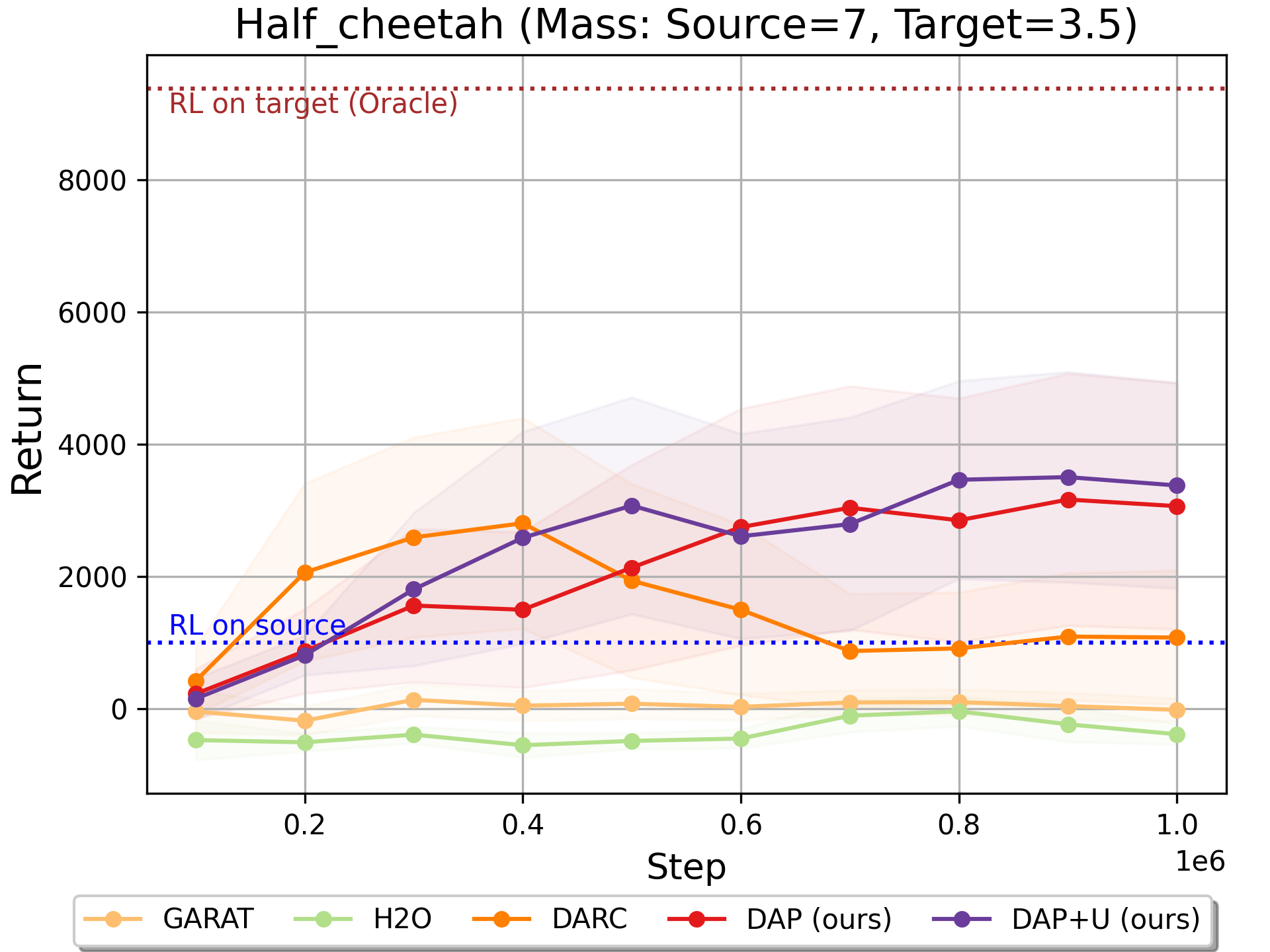}
    \end{subfigure}
    \caption{\textbf{Main results evaluated in target environment.} We compare our proposed methods, DAP and DAP+U, against several baseline approaches.}
    \label{fig:main}
\end{figure}

We evaluated baseline policies across four target tasks.
Each evaluation comprised 100 episodes across 3 random seeds per checkpoint.
The evaluation curves are presented in Figure \ref{fig:main}, with the title specifying details about the tasks and environment settings. We plot target environment returns against the number of source environment training steps.
For ``RL on source'' and ``RL on target", we show their respective maximum returns as horizontal dashed lines. For all tasks, ``RL on source'' performs significantly worse than ``RL on target''. The huge gaps suggest that policies optimized for the source domain are not directly transferable to the target domain.

Our experiments reveal several key findings. First, H2O performs poorly in all tasks, often worse than ``RL on source''. This is because H2O relies on access to optimal target data for good performance. Next, without online interaction, both DARC and GARAT struggle, achieving performance slightly better or worse than ``RL on source.'' Our proposed method, DAP, outperforms all other baselines significantly. It achieves at least double the return compared to ``RL on source'' on all tasks, demonstrating its effectiveness despite utilizing only sub-optimal offline data. Furthermore, DAP with uncertainty estimation (DAP+U) shows additional improvement on most tasks, nearly matching the performance of ``RL on target'' in some cases. This demonstrates the effectiveness of the robust action resampling. Notably, the half-cheetah task remains the most challenging, with even the best method achieving only $\sim3500$, significantly lower than the optimal score of 9000.

\begin{figure}[t]
    \centering
    \begin{subfigure}{0.45\textwidth}
        \centering
        \includegraphics[width=\linewidth]{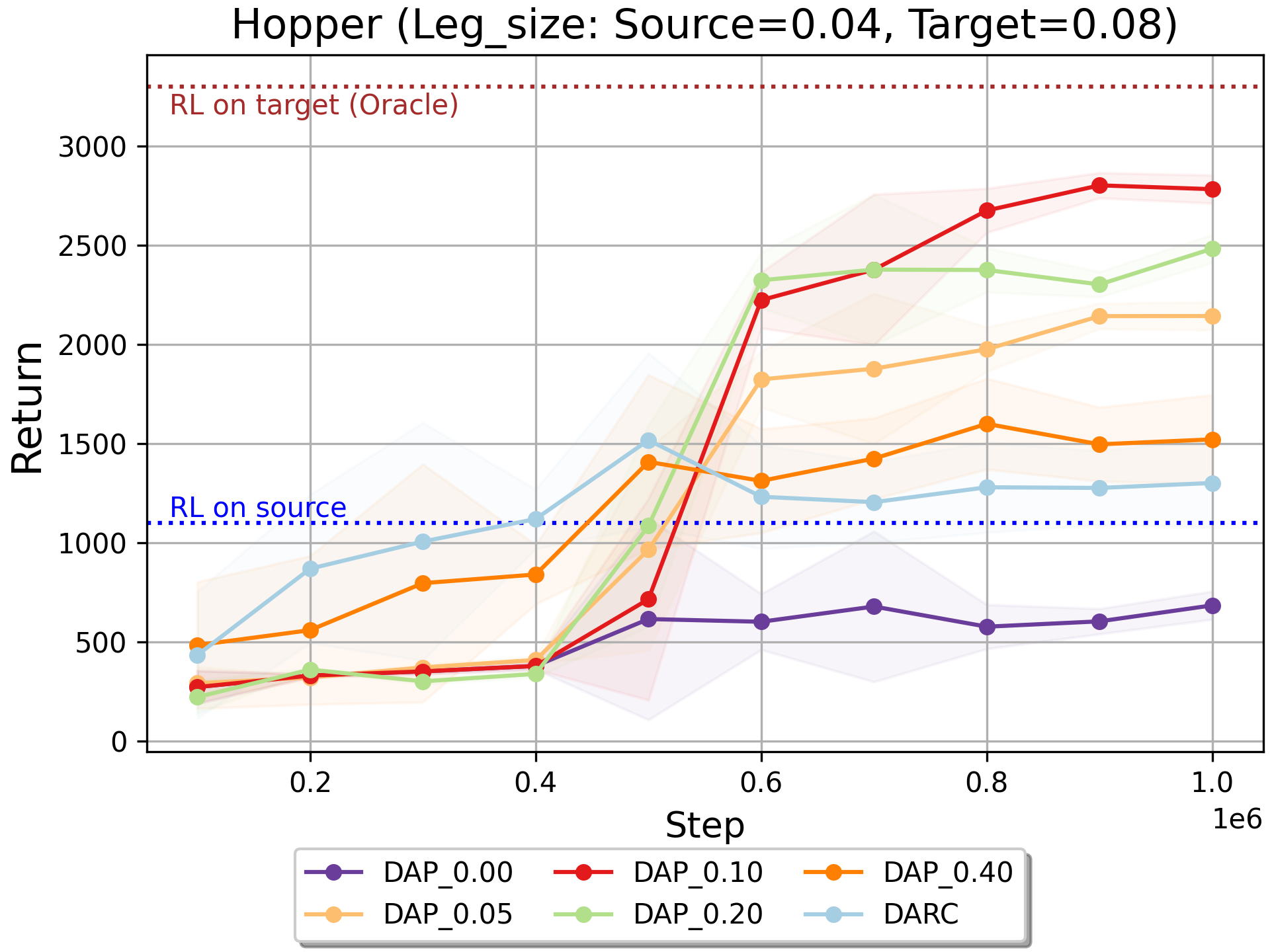}
        \caption{Ablation 1: Regularization Effects on parameter $\lambda$.}
        \label{fig:ab1}
    \end{subfigure}
    \hspace{0.05\textwidth}
    \begin{subfigure}{0.45\textwidth}
        \centering
        \includegraphics[width=\linewidth]{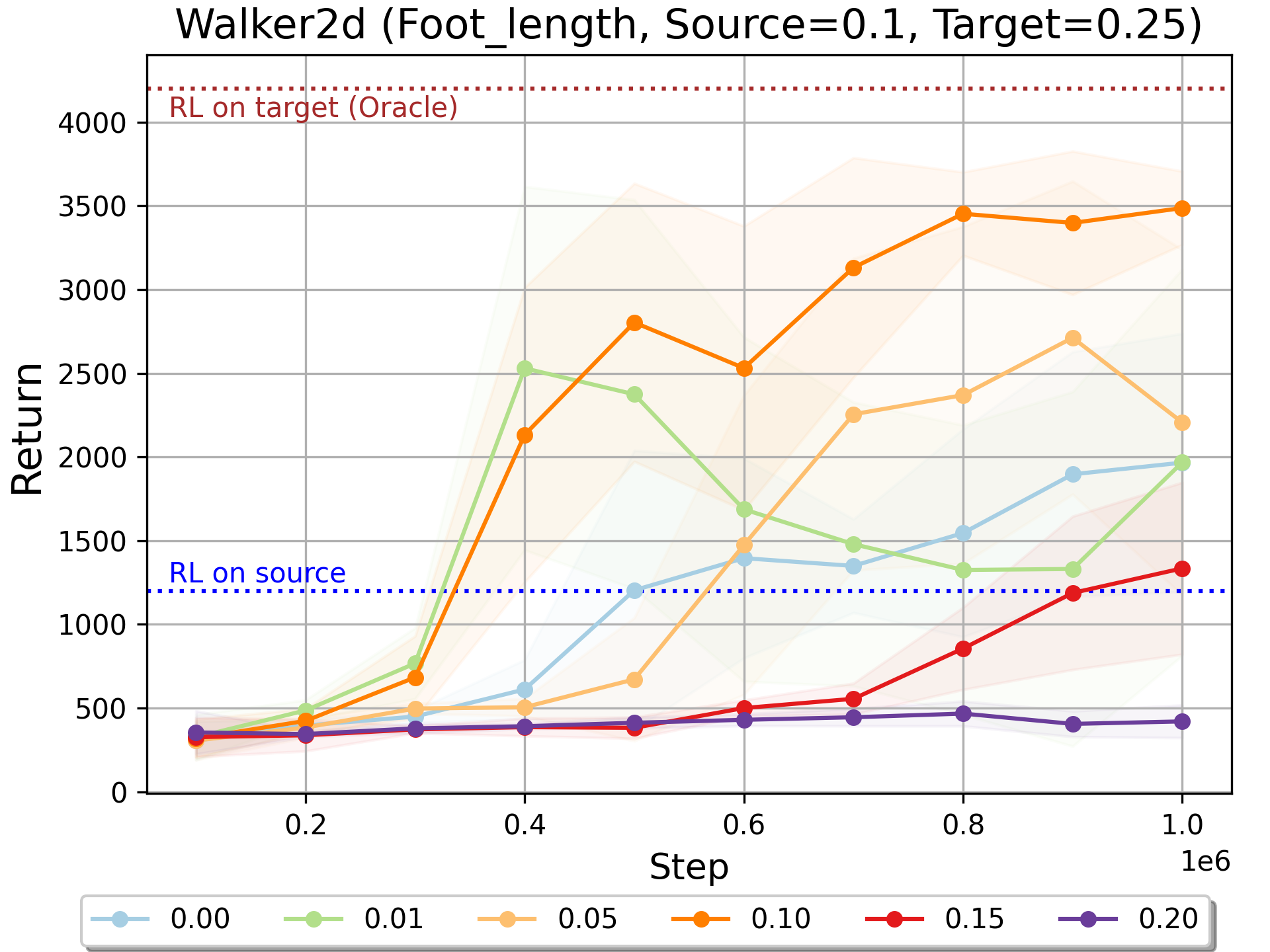}
        \caption{Ablation 2: Effect of scaling parameter for uncertainty-based action resampling, $k$.}
        \label{fig:ab2}
    \end{subfigure}
    \\
    \begin{subfigure}{0.45\textwidth}
        \centering
        \includegraphics[width=\linewidth]{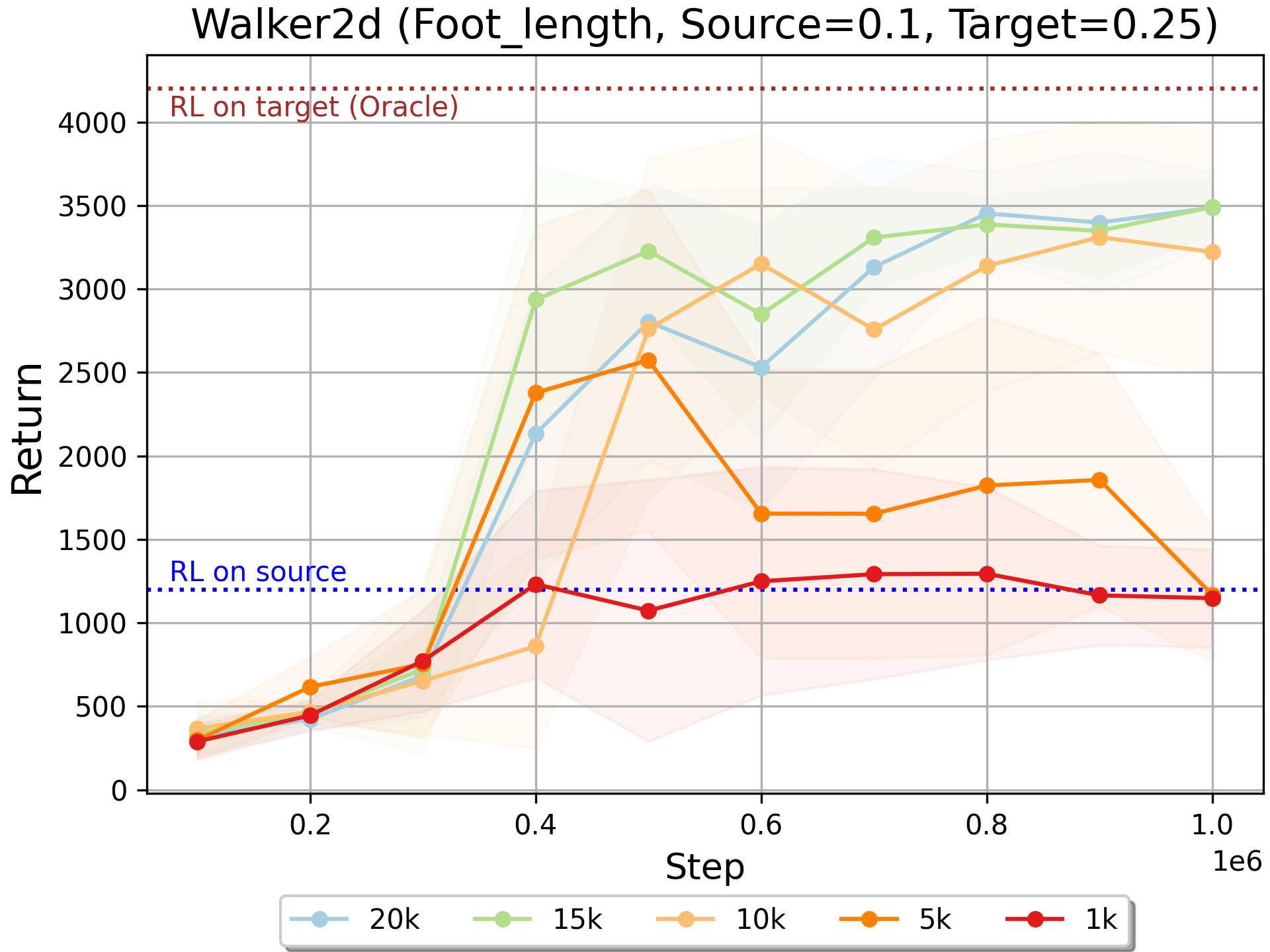}
        \caption{Ablation 3: Effect of size of offline dataset, $M$, collected in target environment.}
        \label{fig:ab3}
    \end{subfigure}
    \label{fig:ablation}
\end{figure}

\subsection{Ablation Experiments}

\textbf{Regularization Effects.} 
Our ablation studies on $\lambda$ as seen in Figure~\ref{fig:ab1} revealed a clear trade-off: as $\lambda$ approaches 0, performance on the target task significantly deteriorates due to the target policy generating "interesting" actions without optimizing returns. Conversely, large $\lambda$ values cause DAP to converge towards DARC-like behavior, as the regularization term dominates the shaping. We found an optimal range for $\lambda$ that preserves DAP's unique benefits while ensuring sufficient closeness between the policies. This balance is critical: it allows the target policy to explore potentially beneficial actions while still leveraging the source policy's optimized behavior. The careful tuning of $\lambda$ thus enables DAP to outperform both unconstrained exploration and strict imitation of the source policy.
\\

\noindent\textbf{Uncertainty-based action resampling.} Our next experiment investigates the importance of the scaling parameter, $k$, for uncertainty-based action resampling (Equation~\ref{eqn:robust}). We evaluate a range of $k$ values on the Walker2D environment and present the results in Figure~\ref{fig:ab2}. We begin with DAP without any resampling ($k=0$) as the baseline. When we slightly increase $k$ to 0.01, we observe slight improvement over the baseline at certain points during training. However, this performance gain doesn't persist throughout the training process. Further increasing $k$ to 0.05 and 0.10 leads to significant improvement over the baseline, with $k=0.10$ achieving the best performance. However, setting $k$ to larger values like 0.15 and 0.20 results in a deterioration of the policy's performance. Notably, at $k=0.20$, the policy exhibits instability and fails completely.
\\

\noindent\textbf{Size of target dataset.}  Our next experiment investigates the impact of the target dataset size, $M$. We evaluate a range of $M$ values on the Walker2D environment and present the results in Figure~\ref{fig:ab3}. The policy setting is DAP+U (DAP with uncertainty estimation, $k=0.10$). We begin with $M=20{,}000$ as the baseline similarly used in the main experiment. We observe that decreasing $M$ to 15{,}000 and 10{,}000 has minimal impact on the results, with the evaluation curves remaining relatively similar throughout the training process. However, a further decrease to $M=5{,}000$ shows a decline in performance beyond 400{,}000 steps. Interestingly, with a bare minimum of $M=1{,}000$ samples (equivalent to 1 episode), the policy maintains performance close to "RL on source" and avoids complete failure.

\section{Conclusion}
This paper introduces Dual Action Policy (DAP), a reinforcement learning method addressing the reality gap. DAP utilizes a single policy to predict two action sets: one maximizing task rewards, the other addressing dynamics mismatch. This decoupling makes it easier to maximise the overall reward in the source domain during training. Experiments demonstrate DAP's superior performance compared to strong baselines, with further improvements observed when incorporating uncertainty estimation.

\section*{ACKNOWLEDGMENT}
This study is supported under RIE2020 Industry Alignment Fund – Industry Collaboration Projects (IAF-ICP) Funding Initiative, as well as cash and in-kind contribution from the industry partner(s). 
%
%
%
%

\bibliographystyle{splncs04}
\bibliography{reference}

\end{document}